\documentclass[
]{ceurart}

\usepackage{listings}
\usepackage{booktabs}
\usepackage{graphicx}
\usepackage{subcaption}
\usepackage{hyperref}
\usepackage{wrapfig}
\usepackage{tcolorbox}
\usepackage{xcolor}
\begin{document}

\copyrightyear{2025}
\copyrightclause{Copyright for this paper by its authors. Use permitted under Creative Commons License Attribution 4.0 International (CC BY 4.0).}

\conference{SymGenAI4Sci 2025: First International Workshop on Symbolic and Generative AI for Science co-located with Semantics-2025, September 3–5, 2025, Vienna, Austria}

\title{Investigating Symbolic Triggers of Hallucination in Gemma Models Across HaluEval and TruthfulQA}

\author[1]{Naveen Lamba}[%
orcid=,
email=naveenlamba30894@gmail.com,
url=,
]
\fnmark[1]
\address[1]{Center for Artificial Intelligence in Medicine, Imaging and Forensics, Sharda University, Greater Noida, India}

\author[1]{Sanju Tiwari}[%
orcid=,
email=tiwarisanju18@ieee.org,
url=,
]
\fnmark[1]
\address[1]{Center for Artificial Intelligence in Medicine, Imaging and Forensics, Sharda University, Greater Noida, India}

\author[2]{Manas Gaur}[%
orcid=,
email=manas@umbc.edu,
url=http://conceptbase.sourceforge.net/mjf/,
]
\fnmark[2]
\address[2]{University of Maryland, Baltimore County, Baltimore, MD, USA}

\fntext[1]{These authors contributed equally.}

\begin{abstract}
   Hallucination in Large Language Models(LLMs) is a well studied problem. However, the properties that make LLM intrinsically vulnerable to hallucinations have not been identified and studied. This research identifies and characterizes the key properties, allowing us to pinpoint vulnerabilities within the model’s internal mechanisms. To solidify on these properties, we utilized two established datasets, HaluEval and TruthfulQA and convert their existing format of question answering into various other formats to narrow down these properties as the reason for the hallucinations. Our findings reveal that hallucination percentages across symbolic properties are notably high for Gemma-2-2B, averaging 79.0\% across tasks and datasets. With increased model scale, hallucination drops to 73.6\% for Gemma-2-9B and 63.9\% for Gemma-2-27B, reflecting a 15 percentage point reduction overall.Although the hallucination rate decreases as the model size increases, a substantial amount of hallucination caused by symbolic properties still persists. This is especially evident for modifiers (ranging from 84.76\% to 94.98\%) and named entities (ranging from 83.87\% to 93.96\%) across all Gemma models and both datasets. These findings indicate that symbolic elements continue to confuse the models, pointing to a fundamental weakness in how these LLMs process such inputs—regardless of their scale.
\end{abstract}


\maketitle

\section{Introduction}

Large language models (LLMs) have made significant advancements in various natural language understanding and generation tasks, including open-domain question answering \cite{Kamalloo2023}, text summarization \cite{Veen2023}, reasoning \cite{Yugeswardeenoo2024}, and dialogue \cite{Guan2025}. Despite their success, the reliability of LLMs remains a major issue due to hallucination\footnote{We use the word ``hallucination'' to be consistent with the terminology used by NLP community, however, we prefer confabulation/fabrication as the appropriate word.}, which involves confidently generating content that is factually inaccurate or nonsensical text \cite{Maynez2020, govil2025cobias}. 

\begin{figure}[h]
    \centering
    \includegraphics[width=0.6\textwidth]{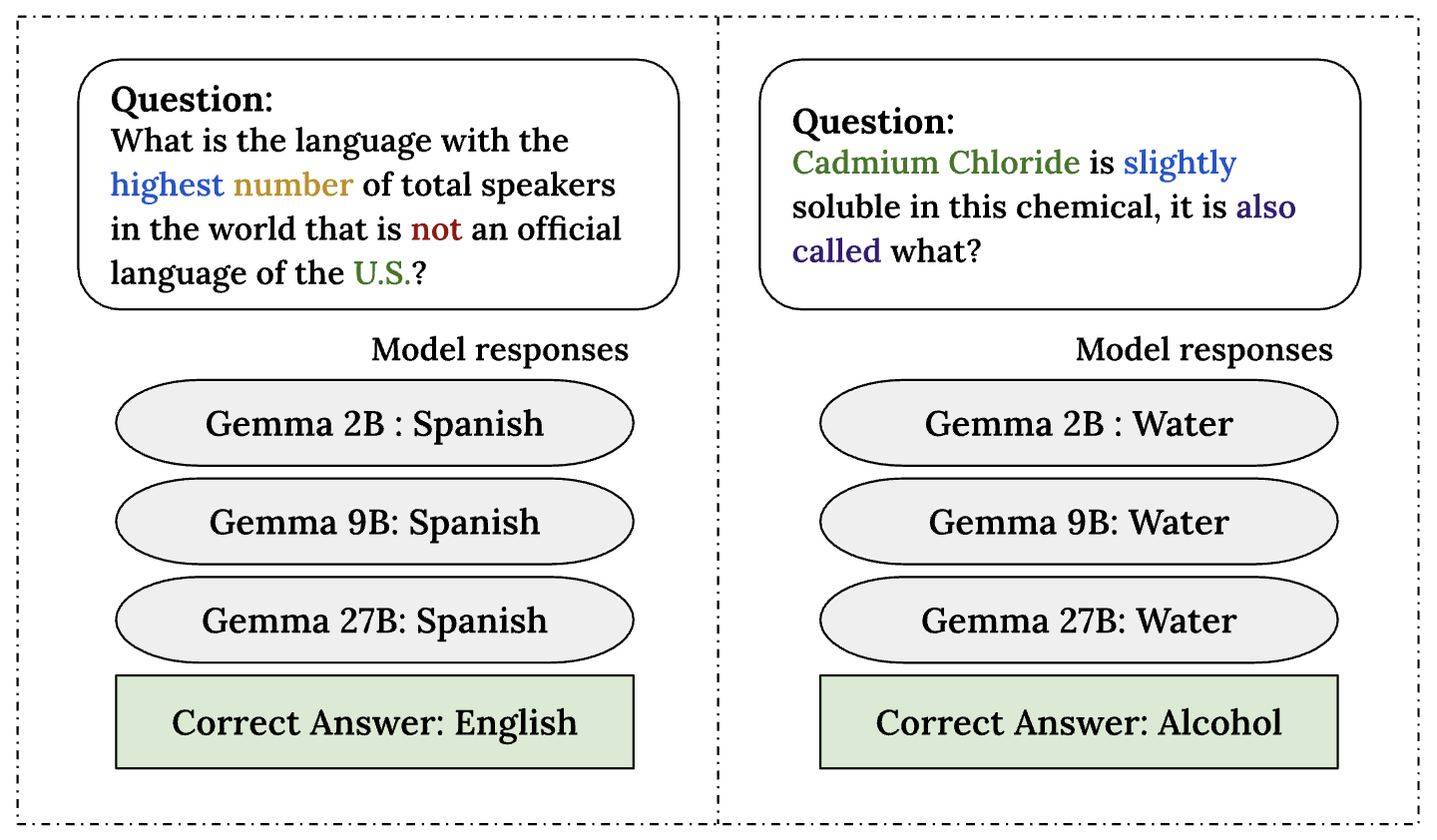}
    \caption{Examples of symbolic triggers causing hallucination across all Gemma model sizes (2B, 9B, 27B). 
    Color coding: \textcolor{blue}{blue} = modifier, \textcolor{red}{red} = negation, \textcolor{violet}{purple} = exception, \textcolor{green}{green} = named entity, \textcolor{yellow!80!orange}{yellow} = number.}
    \label{fig:property}
\end{figure}

While significant research has been conducted on identifying and reducing hallucinations in LLMs \cite{Ji2023, Huang2024}, much of this work has been primarily driven by the development of novel hallucination benchmarks and their corresponding detection and mitigation approaches \cite{Sun2024}. However, the investigation into the fundamental, intrinsic causes of hallucination phenomena in LLMs remains significantly underexplored. Understanding these root causes is particularly crucial because they often stem from limitations in symbolic knowledge representation and reasoning—areas where the NLP community has extensive expertise \cite{Acharya2024, Weston2015}. These limitations manifest through specific and elemental symbolic triggers that consistently provoke hallucinations: \textit{named entities, negation handling, exception cases}, and others can cause LLMs to generate incorrect information, irrespective of the dataset format or domain. \autoref{fig:property} illustrates two such examples where all Gemma models hallucinate in the presence of symbolic triggers like modifiers, named entity, number, negation, and exception. By focusing on these intrinsic mechanisms, researchers can develop more robust, data-agnostic methodologies that not only localize the sources of hallucination within LLMs but also provide systematic, long-term solutions rather than superficial fixes. This deeper understanding would enable the creation of more reliable language models that can better distinguish between accurate and inaccurate generated content, ultimately leading to more trustworthy AI systems \cite{Zhang2023}.

This paper addresses this gap by identifying and describing \textit{symbolic and interpretable} knowledge properties that \textit{reliably trigger hallucination} across natural language understanding task types and model scales. This paper makes several key contributions which are outlined below:  
\paragraph{Key Contributions:}
\begin{itemize}
    \item \textit{Identification of symbolic hallucination triggers:} Systematically identified and characterized five symbolic knowledge properties that reliably trigger hallucination: modifiers (adjectives, adverbs, verbs), named entities, numbers, negation, and exceptions, and provided a property-focused evaluation for understanding intrinsic vulnerabilities in LLMs.
    \item \textit{Prompt engineering-driven data transformation for generalization of symbolic triggers}: Developed a systematic evaluation approach that tests hallucination consistency across three critical dimensions: model scale (Gemma-2-2B, 9B, 27B), task formats (question-answering (QA), multiple choice questions (MCQ), Odd-One-Out (OOO)), and symbolic property types by converting existing datasets to isolate specific triggers, which demonstrates that symbolic vulnerabilities are fundamental architectural issues rather than artifacts of specific experimental conditions.
    \item \textit{Internal activation analysis using symbolic triggers}: Conducted attention pattern analysis and activation-level traces to examine how symbolic properties affect internal model representations and processing, providing evidence that hallucinations stem from deeper representational instabilities rather than surface-level generation errors.
\end{itemize}

These contributions led us to the following findings: 
\begin{itemize}
\item Symbolic triggers elicit hallucination across model sizes: Hallucination rates remain substantially high across all model sizes: 79.0\% (Gemma-2-2B), 73.6\% (Gemma-2-9B), and 63.9\% (Gemma-2-27B), with only a modest 15 percentage point reduction despite significant model scaling, indicating that these are structural rather than capacity-related issues that challenge the assumption that larger models automatically become more reliable. 
\item Primary symbolic triggers: Modifiers show hallucination rates ranging from 84.76\% to 94.98\% across all models while named entities exhibit similarly high rates (83.87\% to 93.96\%), consistently emerging as the most problematic symbolic properties and revealing specific linguistic elements that pose the most significant risk for factual accuracy in LLM outputs.
\item Task Format Dependency: QA format produces the highest hallucination rates compared to MCQ and Odd-One-Out formats, with lower symbolic attention values correlating with higher hallucination frequency, particularly evident in MCQ tasks, demonstrating that task structure significantly influences model reliability and suggesting that constrained generation formats may offer some protection against symbolic confusion.
\item Non-monotonic input length effects of symbolic triggers: Symbolic triggers behave differently across input lengths: modifiers and named entities cause the most hallucinations in short-to-medium contexts (10-30 tokens) but become more reliable with longer context, while numbers follow an unpredictable up-and-down pattern, and negation and exceptions consistently cause fewer problems overall, demonstrating that context length affects each symbolic property uniquely.
\end{itemize}

\section{Related Work}

Recent advances in large language models (LLMs) have intensified focus on understanding and mitigating hallucination—confident outputs that are factually incorrect or logically incoherent. While early research primarily concentrated on output-level detection and dataset-based evaluation of hallucination phenomena\cite{Zhao2020, Durmus2020}, the deeper representational vulnerabilities of LLMs remain underexplored. Our study contributes by examining how hallucination manifests across model sizes, under different task formats, and in response to symbolic properties embedded in inputs.

While a growing body of work evaluates LLMs for factual reliability, few studies assess how hallucination trends evolve with model scale. Notably, works like \citet{Yao2023} frame hallucinations as emergent adversarial phenomena—linked to overconfident generalizations—but do not analyze whether such tendencies vary with parameter count. Similarly, most hallucination benchmarks focus on a single model instance rather than conducting comparative analysis across multiple versions of the same model family. Our work addresses this gap by systematically evaluating hallucination behavior across Gemma-2-2B, 9B, and 27B, revealing that while hallucination rates reduce with scale, symbolic triggers remain persistent.

Benchmark datasets such as TruthfulQA\cite{Lin2021} and HaluEval\cite{Li2023} have been instrumental in evaluating LLM hallucinations. These benchmarks typically use open-ended QA to elicit model generations under minimal constraints, which often reveal factual inconsistencies. However, prior work does not systematically vary task formats to study how structural differences—like constrained generation in multiple-choice or odd-one-out tasks—modulate hallucination tendencies. Our study introduces task format as a key dimension, converting QA data into MCQ and OOO formats to probe whether and how task structure interacts with hallucination triggers.

Numerous research efforts have investigated linguistic and stylistic elements that affect hallucination. For instance, \citet{Rawte2023} demonstrate that the likelihood of hallucinated outputs is influenced by readability, formality, and concreteness. Some have concentrated on particular symbolic structures. Negation has specifically been recognized as a continual vulnerability for LLMs, as shown by \citet{Varshney2024} and \citet{Asher2024}, who illustrate that models often generate false information even when negation indicators are straightforward in syntax and clear in logic. These results indicate that symbolic reasoning continues to be difficult, even if many assessments are limited to individual signals. Our work broadens this scope by evaluating five symbolic properties—modifiers, named entities, numbers, negation, and exceptions—as systematic triggers of hallucination. We also extend analysis beyond surface-level generations, examining how symbolic inputs induce representational instability across transformer layers.

In contrast to prior research, which often isolates one axis of hallucination (model, task, or linguistic feature), our work offers a three-dimensional assessment across model scale, task format, and symbolic input structure. We analyze symbolic hallucination in three Gemma models (2B, 9B, 27B), three reformatted task environments (QA, MCQ, OOO), and five symbolic property types, providing both quantitative trends and internal activation-level insights \cite{joshi2024towards}. This integrative approach reveals that hallucinations are not just artifacts of generation, but reflect deeper weaknesses in how LLMs process structurally complex or logically nuanced inputs.

\section{Methodology}
Our approach involved taking existing datasets, converting them into different question formats, and then testing how three different sizes of Gemma models (2B, 9B, and 27B) responded to questions containing specific symbolic triggers like modifiers, numbers, and named entities.
This study evaluates different versions of Gemma models, open-source checkpoints released by Google DeepMind \cite{Team2024, Team2025}. For consistency across all experiments and to minimize sampling parameter variability, we utilize each model's default temperature value, as provided by the model, which is typically a low or zero value for deterministic generation. This enables us to see the inherent behavior of each model in its recommended decoding setup without injecting sampling-originating randomness.

This research explores the inherent symbolic knowledge characteristics that induce hallucinations in LLMs, i.e., between varying instances of the Gemma model family (2B, 9B, and 27B). The approach follows a systematic, property-focused evaluation pipeline consisting of dataset setup, input transformation, model selection, controlled prompt creation, and hallucination analysis. We base our methodology on the assumption that some input symbolic structures — like modifiers, named entities, negations, numbers, and exceptions — increase the likelihood of LLMs to hallucinate. To empirically test this, we reformatted typical datasets into task-specific ones and inspected the derived outputs.

\begin{figure}
    \centering
    \includegraphics[width=\textwidth]{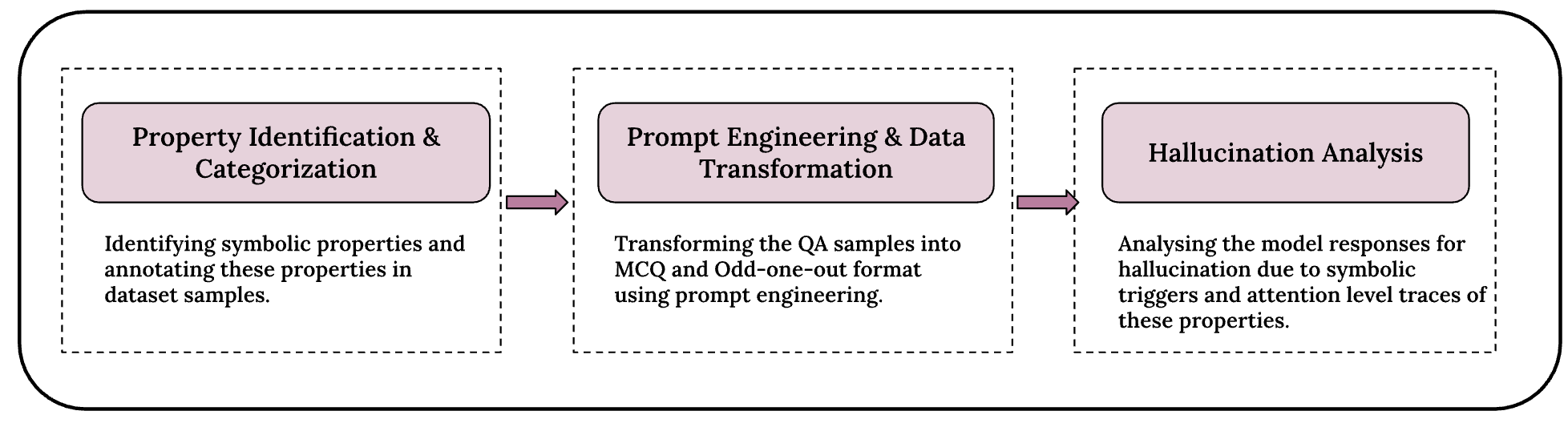}
    \caption{Methodology followed for the hallucination analysis in this research}
    \label{fig:methodology}
\end{figure}


\subsection{Dataset Preparation and Task Conversion}

We used two established hallucination evaluation datasets, HaluEval and TruthfulQA, which contain factual question-answer pairs. To determine whether symbolic triggers cause hallucinations regardless of task structure or if certain formats offer protection against symbolic confusion, we systematically converted these datasets into three distinct formats that provide different levels of generative constraints and cognitive demands: (i) QA format preserves open-ended generation that \textit{may} expose maximum hallucination tendencies since models can freely fabricate plausible-sounding but incorrect responses when encountering symbolic triggers, (ii) MCQ format provides constrained multiple-choice selection that tests whether limiting response options can mitigate symbolic trigger effects by preventing free-form generation, and (iii) OOO format tests semantic classification abilities under symbolic influence to determine if symbolic triggers disrupt fundamental reasoning processes beyond just factual recall.
What we prepared: We systematically transformed 100 samples from each dataset (verified to contain one or more target symbolic properties) into all three task formats, creating a comprehensive evaluation framework of 600 total test instances. Each transformation maintained the core symbolic elements while adapting the response structure to isolate whether symbolic confusion persists across different cognitive demands and constraint levels.
How transformation was achieved: We designed standardized prompts for each format:
QA Prompt: "Answer the following question in one short, factual sentence."
MCQ Prompt: "Consider the following multiple-choice question. Pick the correct answer and explain your reasoning."
Odd One Out Prompt: "Identify the item that does not belong in the list. Explain your reasoning."

For example:
\begin{itemize}
    \item \textbf{QA Prompt: }
    \begin{tcolorbox}[boxrule=0.3pt, colback=gray!5!white, colframe=gray!60!black, left=2pt, right=2pt, top=1pt, bottom=1pt]
    "Answer the following question in one short, factual sentence."
    \end{tcolorbox}

    \item \textbf{MCQ Prompt: }
    \begin{tcolorbox}[boxrule=0.3pt, colback=gray!5!white, colframe=gray!60!black, left=2pt, right=2pt, top=1pt, bottom=1pt]
    "Consider the following multiple-choice question. Pick the correct answer and explain your reasoning."
    \end{tcolorbox}

    \item \textbf{Odd One Out Prompt: }
    \begin{tcolorbox}[boxrule=0.3pt, colback=gray!5!white, colframe=gray!60!black, left=2pt, right=2pt, top=1pt, bottom=1pt]
    "Identify the item that does not belong in the list. Explain your reasoning."
    \end{tcolorbox}
\end{itemize}

\begin{figure}[t]
    \centering
    \includegraphics[width=\textwidth]{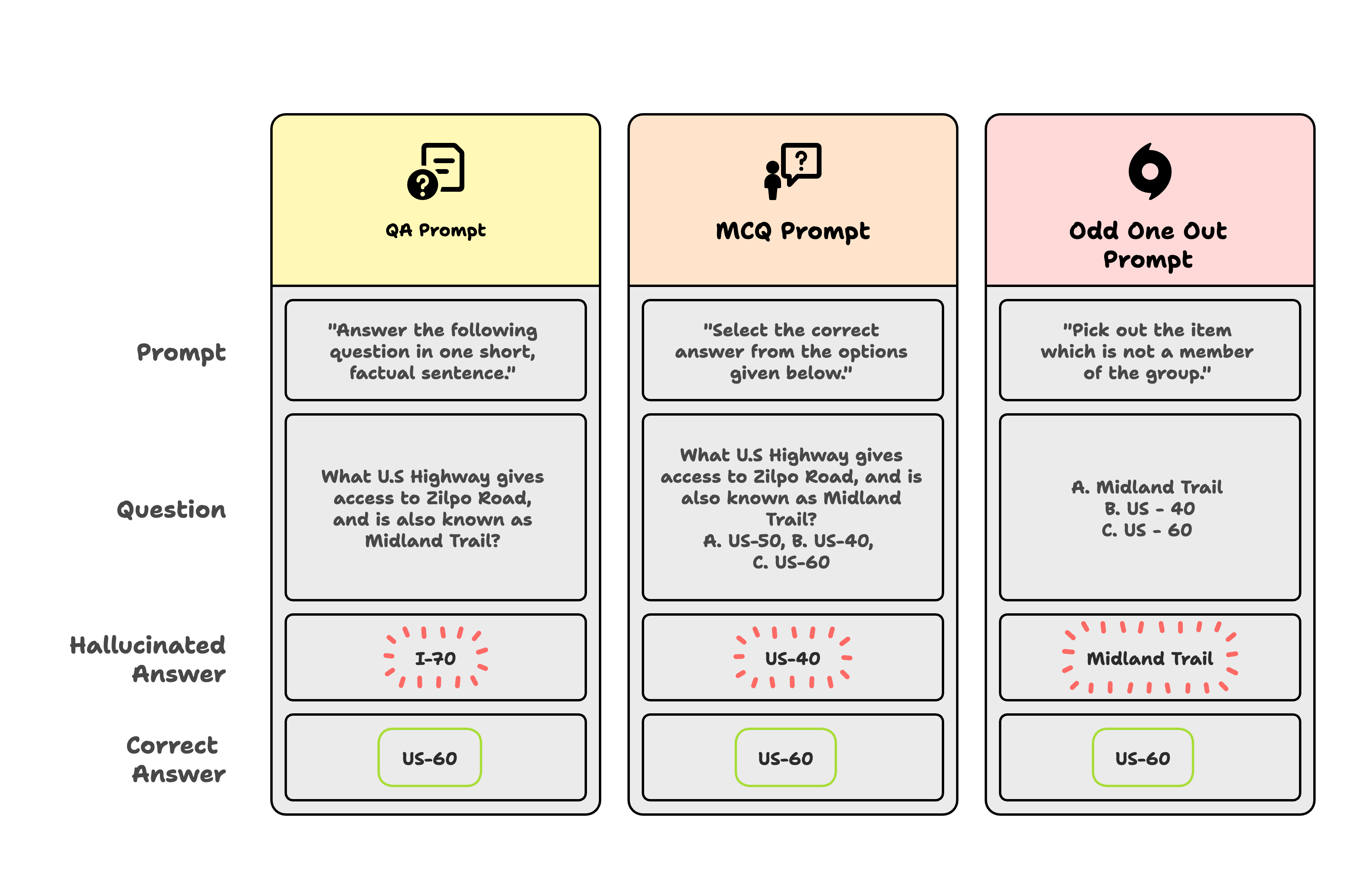}
    \caption{Examples of questions and answers in different formats(QA, MCQ and Odd-One-Out), generated from Gemma-2-2b.}
    \label{fig:enter-label}
\end{figure}
Prompt design was carefully managed so that hallucinations, when they occur, can be attributed to model reasoning and symbolic processing rather than prompt ambiguity, with all transformed prompts annotated for symbolic property analysis.

Two standard hallucination benchmarking datasets: HaluEval\cite{Li2023} and TruthfulQA\cite{Lin2021} have been used for this study. Both datasets include factual QA pairs. To investigate hallucination behavior with varying knowledge forms, we design three task formulations: (i) QA: Preserves the question-answer pair format of the original dataset. (ii) MCQ: Converts every QA pair to a single correct and two distractor options multiple-choice format. (iii) Odd-One-Out: Presents conceptually related options with the exception of one, seeking identification of the semantic outlier.
All 600 samples of the two datasets are checked to have one or more of the five symbolic knowledge properties.

\subsection{Property Identification and Categorization}
To investigate symbolic triggers of hallucination, all input prompts were annotated for the presence of five key symbolic properties. These properties were chosen based on their structural role in language. Each property was identified using linguistic markers and then manually verified to ensure semantic relevance. Here, we define each category, provide examples, and summarize their contribution to hallucination:
\begin{enumerate}
    \item \textbf{Modifiers(adjectives, adverbs, and verbs):} These elements introduce subjective or descriptive information, often adding interpretive flexibility. \textit{Example: “Which is the most rapidly growing city in Europe?”}. Modifiers such as “rapidly” or “most” invite vague or ambiguous completions, increasing the risk of confident but unverifiable assertions. LLMs may hallucinate plausible-sounding answers even when the modifier-driven nuance is not grounded in training data.
    \item \textbf{Named Entities(persons, organizations, locations):} Identified using Named Entity Recognition (NER) techniques, these refer to proper nouns that often require external knowledge grounding. \textit{Example: “Who is the founder of the fictional company TechNova?”}. Due to their reliance on memorized or incomplete knowledge, LLMs often fabricate facts or assign incorrect associations when dealing with named entities—especially rare or fictional ones.
    \item \textbf{Numbers(quantitative expressions):} These include cardinal numbers, ranges, dates, and measurements. \textit{Example: “How many satellites does Mars currently have?”}. LLMs are prone to imprecision or outright numerical hallucination, either due to outdated training data or due to overgeneralizing learned patterns. Such prompts demand factual accuracy, making errors more noticeable.
    \item \textbf{Negation(not, never, none, cannot):} Detected via syntactic and semantic analysis, negation alters the logical polarity of a sentence. \textit{Example: “Which of these is not a fruit?”}. LLMs frequently mishandle negation by overlooking or misinterpreting the negative cue, resulting in logically inverted or irrelevant answers.
    \item \textbf{Exceptions(edge cases, conditional rules):} These refer to inputs that challenge the model to recognize rare or counterexamples. \textit{Example: “Which metal is liquid at room temperature?”}. Exceptions require deeper contextual reasoning. Since LLMs tend to generalize, they often miss these special cases, favoring the more common rule rather than the exception.
\end{enumerate}

By categorizing prompts along these symbolic dimensions, we aim to isolate specific triggers that systematically increase hallucination likelihood across tasks and model scales. This property-level lens provides a more interpretable understanding of why and when LLMs go wrong.




\subsection{Hallucination Evaluation Strategy}
We employ a three-tier hallucination analysis approach, progressing from overall hallucination rates to detailed, layer-wise causes, ultimately attributing them to five symbolic triggers.

\noindent \textit{Symbolic trigger-based computation of hallucination percentage:} To quantify hallucination induced by symbolic properties, we annotated each input for the presence of one or more symbolic triggers (modifiers, named entities, numbers, negation, exceptions) and computed the proportion of hallucinated outputs within each trigger category. A prediction was marked as a hallucination if it was factually incorrect. This computation was carried out per symbolic property, allowing us to isolate their individual contribution to hallucination rates. The final hallucination percentage per property was then calculated as the number of hallucinated instances containing that property divided by the total instances containing it.

\noindent \textit{Symbolic trigger-driven attention analysis of Gemma models:} We analyze attention scores to symbolic tokens at specific transformer layers selected based on prior research patterns. Following \citet{Wu2025}'s approach, which emphasizes mid-to-deeper layers where semantic integration peaks, we examine Layers 10 and 20 for Gemma-2-2B, Layers 20 and 31 for Gemma-2-9B, and Layers 23 and 40 for Gemma-2-27B. This allows consistent comparison of symbolic attention allocation across model sizes.

\noindent \textit{Input token length and hallucination percentage analysis:} We investigate the relationship between hallucination rates and input question length by organizing data into token length bins and analyzing how symbolic property effects vary across different context sizes. This reveals whether symbolic triggers have consistent effects regardless of the surrounding context or if their impact changes with input complexity.

\section{Results and Analysis}

This section presents our empirical analysis of hallucination behavior in the Gemma model family under symbolic property influence. The investigation is organized along three axes: (i) consistency across model sizes, (ii) variation across task types, and (iii) internal activation responses. The evaluation spans all five symbolic property types, with hallucination annotated as confident yet factually incorrect responses.

\subsection{Consistency Across Model Variants}
In the QA format, modifiers, named entities, and numbers consistently emerge as the most hallucination-prone symbolic properties across all three Gemma model sizes. As shown in \autoref{tab:combined_QA_symbolic}, hallucination percentages for modifiers in the HaluEval dataset remain notably high, decreasing only slightly from 84.76\% in Gemma-2-2B to 77.24\% in Gemma-2-27B. Named entities follow a similar trend, with a marginal drop from 83.87\% to 76.43\%, while numbers stay persistently high at around 83.16\%–76.32\% across model scales.

This pattern is also observed in the TruthfulQA dataset, where modifiers reach up to 94.98\% in Gemma-2-9B and numbers peak at 98.00\%, reflecting the models' continued struggle with these symbolic cues. On the other hand, while negation and exceptions appear less frequently in HaluEval (e.g., 70.00\% and 80.00\% in Gemma-2-2B), their hallucination rates remain above 90\% in TruthfulQA across all model sizes.

These results indicate that scaling up model size offers only modest reductions in hallucination rates for symbolic properties, and that the same set of symbolic triggers continues to challenge LLMs, revealing a persistent internal vulnerability.

\begin{table}[h]
\centering
\caption{Symbolic hallucination percentage statistics for QA task across model sizes and datasets.}
\scriptsize
\begin{tabular}{l|cc|cc|cc}
\toprule
\multirow{2}{*}{\textbf{Symbolic Property}} & \multicolumn{2}{c|}{\textbf{Gemma-2-2B}} & \multicolumn{2}{c|}{\textbf{Gemma-2-9B}} & \multicolumn{2}{c}{\textbf{Gemma-2-27B}} \\
 & HaluEval & TruthfulQA & HaluEval & TruthfulQA & HaluEval & TruthfulQA \\
\midrule
Modifiers         & 84.76 & 89.12  & 77.45 & 94.98 & 77.24 & 86.19 \\
Named Entities    & 83.87 & 89.01 & 77.17 & 93.96 & 76.43 & 88.46 \\
Numbers           & 83.16 & 96.00 & 75.26 & 98.00 & 76.32 & 94.00 \\
Negation          & 70.00 & 91.67 & 70.00 & 95.83 & 80.00 & 95.83 \\
Exceptions        & 100.0 & 94.44 & 80.00 & 96.30 & 80.00 & 90.74 \\
\bottomrule
\end{tabular}
\label{tab:combined_QA_symbolic}
\end{table}

\subsection{Generalization Across Task Formats}
To understand how hallucination behavior generalizes across prompt formats, we analyzed symbolic token attention across QA, MCQ, and Odd-One-Out (OOO) tasks using the Gemma model family (2B, 9B, 27B). \autoref{tab:attention_symbolic} presents average attention scores to symbolic tokens across task formats and model sizes, measured at specific mid-to-deeper layers.

Following prior layer selection patterns used in \citet{Wu2025}, which emphasized mid and post-mid transformer layers ( Layers 10 and 20 for Gemma-2-2B and Layers 20 and 31 for Gemma-2-9B), we chose Layers 23 and 40 for Gemma-2-27B. These lie in the middle-to-late segments of the model, where semantic integration and abstract token interactions typically peak. This alignment allows for a consistent and meaningful comparison of symbolic attention across model sizes.

The results indicate that task format substantially affects both hallucination frequency and attention allocation, despite using prompts with similar symbolic triggers. Across all model sizes, MCQ prompts result in consistently higher hallucination frequency than QA, particularly at the 2B scale. This correlates with lower symbolic attention values for MCQ compared to QA—suggesting reduced grounding or interpretive focus. For instance, in the 27B model, attention to modifiers in QA is 0.0078 (Layer 23), dropping to 0.0063 in MCQ, and further varying in OOO (0.0085). This indicates task-specific shifts in symbolic emphasis, even within the same model.

\begin{table}[h]
\centering
\caption{Attention of symbolic tokens at different layers for QA task across model sizes and datasets.}
\scriptsize
\begin{tabular}{c|l|cc|cc|cc}
\toprule
\multirow{2}{*}{\textbf{Task}} &\multirow{2}{*}{\textbf{Symbolic Property}} & \multicolumn{2}{c|}{\textbf{Gemma-2-2B}} & \multicolumn{2}{c|}{\textbf{Gemma-2-9B}} & \multicolumn{2}{c}{\textbf{Gemma-2-27B}} \\
 & & Layer 10 & Layer 20 & Layer 20 & Layer 31 & Layer 23 & Layer 40 \\
\midrule
\multirow{5}{*}{QA} & Modifiers         & 0.0100 & 0.0097 & 0.0095 & 0.0092 & 0.0078 & 0.0059 \\
                    & Named Entities    & 0.0147 & 0.0082 & 0.0168 & 0.0095 & 0.0165 & 0.0063 \\
                    & Numbers           & 0.0114 & 0.0056 & 0.0122 & 0.0060 & 0.0117 & 0.0047 \\
                    & Negation          & 0.0172 & 0.0091 & 0.0182 & 0.0070 & 0.0137 & 0.0062 \\
                    & Exceptions        & 0.0166 & 0.0118 & 0.0134 & 0.0101 & 0.0158 & 0.0072 \\
\midrule

\multirow{5}{*}{MCQ}& Modifiers         & 0.0093 & 0.0068 & 0.0084 & 0.0067 & 0.0063 & 0.0039 \\
                    & Named Entities    & 0.0177 & 0.0051 & 0.0134 & 0.0062 & 0.0116 & 0.0040 \\
                    & Numbers           & 0.0104 & 0.0038 & 0.0095 & 0.0043 & 0.0085 & 0.0030 \\
                    & Negation          & 0.0206 & 0.0067 & 0.0147 & 0.0052 & 0.0103 & 0.0042 \\
                    & Exceptions        & 0.0140 & 0.0083 & 0.0107 & 0.0072 & 0.0121 & 0.0050 \\
\midrule

\multirow{5}{*}{OOO}& Modifiers         & 0.0076 & 0.0071  & 0.0077 & 0.0062 & 0.0085 & 0.0040 \\
                    & Named Entities    & 0.0087 & 0.0055 & 0.0123 & 0.0055 & 0.0106 & 0.0035 \\
                    & Numbers           & 0.0050 & 0.0031 & 0.0074 & 0.0032 & 0.0063 & 0.0052 \\
                    & Negation          & 0.0092 & 0.0068 & 0.0084 & 0.0054 & 0.0082 & 0.0035 \\
                    & Exceptions        & 0.0072 & 0.0047 & 0.0070 & 0.0064 & 0.0085 & 0.0035 \\
\bottomrule
\end{tabular}
\label{tab:attention_symbolic}
\end{table}

Conversely, while OOO prompts show relatively lower symbolic attention, they elicit stronger semantic hallucination effects, particularly in smaller models (as seen in prior hallucination rate and effect metrics). Notably, in the 2B model, symbolic attention for named entities drops sharply in MCQ (0.0177 → 0.0051 from Layer 10 to 20), whereas QA retains higher symbolic focus (0.0147 → 0.0082). The same trend, though attenuated, persists in 27B, showing a consistent symbolic property ranking: modifiers and named entities receive the highest attention, followed by numbers, negation, and exceptions.

\subsection{Activation-Level Traces of Symbolic Instability}
To further probe the internal behavior of LLMs in the presence of symbolic linguistic properties, we analyzed the relationship between hallucination and input question length. \autoref{tab:symbolic_to_length} illustrate the average hallucination percentages across symbolic properties (modifiers, named entities, numbers, negation, and exceptions) as a function of token length, for both HaluEval and TruthfulQA datasets. We observe that hallucination induced by symbolic properties like modifiers and named entities remains consistently high across varying input lengths. For instance, modifiers peaked at nearly 97\% hallucination in 0–29 query token length bracket, while named entities followed a similar trend with a peak around 78\%, which is actually the normal length of the query used by a layman. Notably, hallucination rates tend to decline for longer queries (40+ tokens), potentially due to enhanced contextual grounding, although the trend is not uniform across all properties. Instances where hallucination percentages drop to 0\% are due to the absence of the corresponding symbolic property in that token-length bracket. However, as evident from the table, even minimal presence of a property often corresponds with noticeable hallucination, underscoring a persistent underlying effect.





\begin{table}[h]
\centering
\caption{Hallucination \% by symbolic property and question token length across Gemma models and datasets (HaluEval, TruthfulQA). For HaluEval, hallucination values for the \textbf{exception} property are 0 in all length bins except 10–19 and 30–39 due to limited occurrences. For TruthfulQA, hallucination \% is 0 in the 50+ length bin, as all questions were shorter than 50 tokens.}

\scriptsize
\begin{tabular}{c|c|cc|cc|cc}
\toprule
\rowcolor{gray!30}\textbf{Query Token} & \textbf{Symbolic} & \multicolumn{2}{c|}{\textbf{2B}} & \multicolumn{2}{c|}{\textbf{9B}} & \multicolumn{2}{c}{\textbf{27B}} \\
 \rowcolor{gray!30}\textbf{Length}& \textbf{Property} & HaluEval & TruthfulQA & HaluEval & TruthfulQA & HaluEval & TruthfulQA \\
\midrule
\multirow{5}{*}{0–9}
 & \cellcolor{gray!30}Modifiers        & 61.40 & 83.59 & 66.67 & 87.89 & 61.40 & 78.91 \\
 & \cellcolor{gray!30}Named Entities   & 47.37 & 25.78 & 49.12 & 25.78 & 45.61 & 25.39 \\
 & \cellcolor{gray!30}Numbers          & 10.53 & 7.42  & 8.77  & 7.03  & 8.77  & 7.03  \\
 & \cellcolor{gray!30}Negation         & 0.00  & 2.34  & 0.00  & 2.73  & 0.00  & 2.34  \\
 & \cellcolor{gray!30}Exceptions       & 0.00  & 7.81  & 0.00  & 8.59  & 0.00  & 7.42  \\
\midrule
\multirow{5}{*}{10–19}
 & \cellcolor{gray!30}Modifiers        & 85.42 & 88.50 & 78.31 & 97.00 & 77.97 & 88.50 \\
 & \cellcolor{gray!30}Named Entities   & 67.80 & 33.50 & 63.73 & 37.50 & 62.71 & 34.50 \\
 & \cellcolor{gray!30}Numbers          & 27.12 & 8.00  & 25.42 & 8.00  & 24.75 & 8.00  \\
 & \cellcolor{gray!30}\cellcolor{gray!30}Negation         & 0.68  & 6.50  & 0.68  & 7.00  & 1.02  & 7.00  \\
 & \cellcolor{gray!30}Exceptions       & 1.02  & 13.50 & 0.68  & 13.50 & 0.68  & 13.00 \\
\midrule
\multirow{5}{*}{20–29}
 & \cellcolor{gray!30}Modifiers        & 84.47 & 87.88 & 72.82 & 87.88 & 76.70 & 81.82 \\
 & \cellcolor{gray!30}Named Entities   & 79.61 & 75.76 & 67.96 & 75.76 & 71.84 & 69.70 \\
 & \cellcolor{gray!30}Numbers          & 50.49 & 30.30 & 43.69 & 33.33 & 46.60 & 30.30 \\
 & \cellcolor{gray!30}Negation         & 1.94  & 9.09  & 1.94  & 6.06  & 1.94  & 9.09  \\
 & \cellcolor{gray!30}\cellcolor{gray!30}Exceptions       & 0.00  & 12.12 & 0.00  & 9.09  & 0.00  & 12.12 \\
\midrule
\multirow{5}{*}{30–39}
 & \cellcolor{gray!30}Modifiers        & 72.41 & 57.14 & 68.97 & 42.86 & 62.07 & 57.14 \\
 & \cellcolor{gray!30}\cellcolor{gray!30}Named Entities   & 65.52 & 42.86 & 62.07 & 42.86 & 55.17 & 42.86 \\
 & \cellcolor{gray!30}Numbers          & 48.28 & 28.57 & 48.28 & 28.57 & 48.28 & 28.57 \\
 & \cellcolor{gray!30}Negation         & 6.90  & 0.00  & 6.90  & 0.00  & 6.90  & 0.00  \\
 & \cellcolor{gray!30}Exceptions       & 6.90  & 0.00  & 6.90  & 0.00  & 6.90  & 0.00  \\
\midrule
\multirow{5}{*}{40–49}
 & \cellcolor{gray!30}\cellcolor{gray!30}Modifiers        & 66.67 & 66.67 & 41.67 & 66.67 & 41.67 & 66.67 \\
 & \cellcolor{gray!30}Named Entities   & 58.33 & 33.33 & 41.67 & 33.33 & 33.33 & 33.33 \\
 & \cellcolor{gray!30}Numbers          & 33.33 & 33.33 & 25.00 & 33.33 & 25.00 & 33.33 \\
 & \cellcolor{gray!30}Negation         & 8.33  & 0.00  & 8.33  & 0.00  & 8.33  & 0.00  \\
 & \cellcolor{gray!30}Exceptions       & 0.00  & 0.00  & 0.00  & 0.00  & 0.00  & 0.00  \\
\midrule
\multirow{5}{*}{50+}
 & \cellcolor{gray!30}Modifiers        & 75.00 & 0.00  & 50.00 & 100.00 & 75.00 & 0.00  \\
 & \cellcolor{gray!30}Named Entities   & 75.00 & 0.00  & 50.00 & 100.00 & 75.00 & 0.00  \\
 & \cellcolor{gray!30}Numbers          & 50.00 & 0.00  & 25.00 & 100.00 & 50.00 & 0.00  \\
 & \cellcolor{gray!30}Negation         & 0.00  & 0.00  & 0.00  & 0.00  & 0.00  & 0.00  \\
 & \cellcolor{gray!30}Exceptions       & 0.00  & 0.00  & 0.00  & 0.00  & 0.00  & 0.00  \\
\bottomrule
\end{tabular}
\label{tab:symbolic_to_length}
\end{table}

These observations suggest that certain symbolic properties evoke unstable internal activations, especially in shorter to mid-length prompts. The model's inability to generalize robustly across symbolic structures, regardless of input size, reveals activation-level fragility tied to linguistic form, rather than token count alone. This provides evidence that hallucinations are not solely a product of length context, but of deeper symbolic entanglement.

Our findings strongly indicate that symbolic linguistic properties, particularly modifiers, named entities, and numbers, act as consistent triggers for hallucination across all Gemma model sizes. While scaling from Gemma-2-2B to 27B reduces hallucination rates modestly (by ~15 percentage points), symbolic hallucinations persist even in the largest models. This persistence highlights that such hallucinations are not solely a function of model capacity but stem from how these models internally encode and generalize over symbolic constructs.
Additionally, our activation-level analysis reveals that hallucination rates vary with input length, peaking for mid-range lengths (10–30 tokens). This suggests that context size interacts nonlinearly with symbolic processing, which may indicate local representational instability rather than mere underfitting. Across all models and tasks, QA emerges as the most hallucination-prone format, reinforcing that generative responses under minimal constraints (unlike MCQ or OOO) expose deeper symbolic weaknesses in LLMs.

\section{Conclusion and Future Directions}

This study presents a focused investigation into the symbolic triggers of hallucination in Gemma language models. Across tasks and datasets, we consistently observe that hallucinations are most frequently associated with symbolic linguistic properties—especially modifiers, named entities, and numbers. While scaling the model from Gemma-2-2B to 27B results in a modest reduction in hallucination rates, these symbolic vulnerabilities persist regardless of model size, revealing a deeper representational fragility. Our activation-level analyses further suggest that hallucination is not merely a product of input length or task format, but is tightly coupled with how LLMs internalize and generalize over symbolic structures. The persistence of high hallucination rates, particularly in QA tasks, indicates that symbolic confusion remains a core limitation of current LLM architectures. However, now we have symbolic knowledge that can help us locate hallucination within the layers of open-source LLMs. 

The future work will focus on two key technical directions: Mechanistic interpretability analysis will employ activation patching and causal intervention techniques to precisely localize which transformer layers and attention heads are responsible for symbolic confusion, enabling targeted architectural improvements. Cross-model generalizability studies will systematically validate these symbolic vulnerabilities across different model families (LLaMA, Mistral, GPT) to determine whether these represent universal architectural limitations or model-specific weaknesses. We also aim to extend this analysis to multilingual and multimodal LLMs to evaluate the generality of symbolic hallucinations across modalities and languages. Finally, exploring prompt-based interventions may offer practical mitigation strategies by reducing symbolic ambiguity at inference time.

\begin{acknowledgments}
  The authors gratefully acknowledge the use of the NVIDIA H100 DGX system provided by the Centre for Artificial Intelligence in Medicine, Imaging and Forensics (CAIMIF), Sharda University, which made the large-scale model experiments and evaluations possible.
\end{acknowledgments}

\section*{Declaration on Generative AI}
During the preparation of this work, the author(s) used ChatGPT for Grammar and spelling check. After using this, the author(s) reviewed and edited the content as needed and take(s) full responsibility for the publication’s content.


\bibliography{sample-1col}
\appendix

\section{Online Resources}

The source code and data related to this work are available at:
\begin{itemize}
\item \href{https://github.com/naveen308/Symbolic_triggers.git}{GitHub}
\end{itemize}

\end{document}